\documentclass[10pt,twocolumn,letterpaper]{article}

\usepackage[pagenumbers]{cvpr} 

\definecolor{cvprblue}{rgb}{0.21,0.49,0.74}
\usepackage[pagebackref,breaklinks,colorlinks,allcolors=cvprblue]{hyperref}
\usepackage{graphicx}
\usepackage{amsmath}
\usepackage{amssymb}
\usepackage{booktabs}
\usepackage{multirow}
\usepackage{floatrow}
\usepackage{tcolorbox}
\usepackage{threeparttable}
\usepackage{color}
\usepackage{soul}

\newcommand{\yx}[1]{{\color{black} #1}}

\usepackage{bm}

\title{Omni-Emotion: Extending Video MLLM with Detailed Face and Audio Modeling for Multimodal Emotion Analysis}

\author{Qize Yang$^{1}$\thanks{: Corresponding authors. Project: \url{https://github.com/HumanMLLM/Omni-Emotion}} \quad Detao Bai$^{1}$ \quad Yi-Xing Peng$^{1,2*}$ \quad Xihan Wei$^{1}$ \\
$^1$Tongyi Lab, Alibaba Group; \qquad$^2$Sun Yat-sen University, China. 
}

\begin{document}
\maketitle
\begin{abstract}
	
Understanding emotions accurately is essential for fields like human-computer interaction. Due to the complexity of emotions and their multi-modal nature (e.g., emotions are influenced by facial expressions and audio), researchers have turned to using multi-modal models to understand human emotions rather than single-modality. However, current video multi-modal large language models (MLLMs) encounter difficulties in effectively integrating audio and identifying subtle facial micro-expressions. Furthermore, the lack of detailed emotion analysis datasets also limits the development of multimodal emotion analysis. To address these issues, we introduce a self-reviewed dataset and a human-reviewed dataset, comprising 24,137 coarse-grained samples and 3,500 manually annotated samples with detailed emotion annotations, respectively. These datasets allow models to learn from diverse scenarios and better generalize to real-world applications. Moreover, in addition to the audio modeling, we propose to explicitly integrate facial encoding models into the existing advanced Video MLLM, enabling the MLLM to effectively unify audio and the subtle facial cues for emotion understanding. By aligning these features within a unified space and employing instruction tuning in our proposed datasets, our Omni-Emotion achieves state-of-the-art performance in both emotion recognition and reasoning tasks.
\end{abstract}

\section{Introduction}
\label{sec:intro}
\begin{figure}[t]
      \centering
      \includegraphics[height=7cm,trim=0 0 0 0,clip]{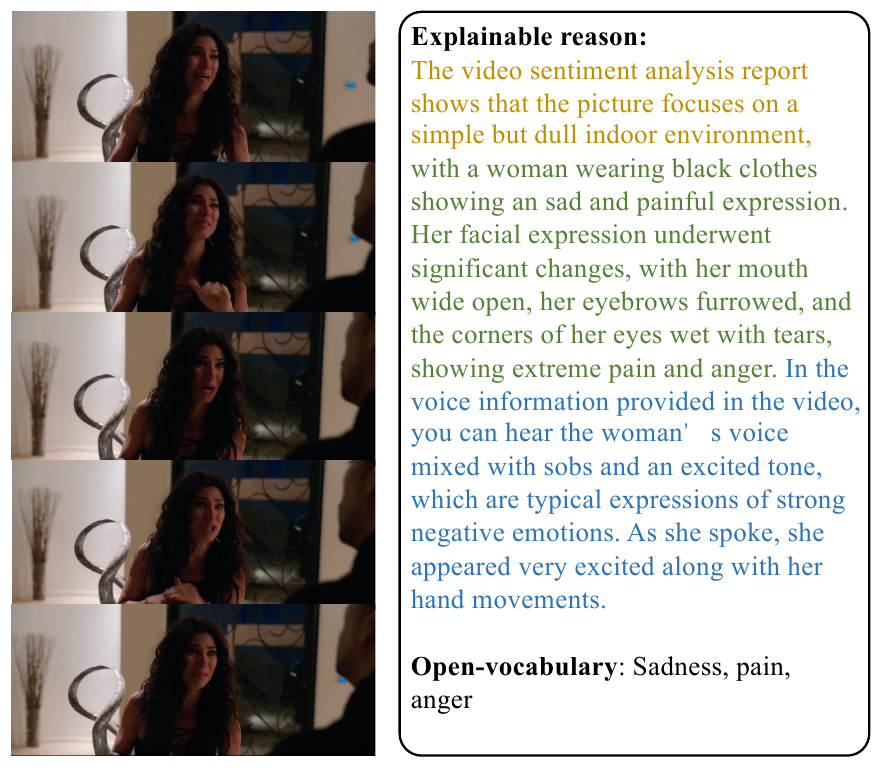}
      \caption{An example from MAFW~\cite{liu2022mafw} with our self-reviewed annotation shows the integration of multimodal explainable reasons for emotion analysis, involving background information, detailed facial expressions, and audio cues. Besides, the open vocabulary labels provide a more accurate description of the character's emotions in the video.
      }
      \label{fig:intro}
\end{figure}

Emotion analysis plays a crucial role in areas such as educational tools, avatars, and human-computer interaction.  Traditional single-modal methods such as audio-based emotion recognition~\cite{fan2021lssed,kondratenko2022large,hsu2021hubert}, text sentiment analysis~\cite{lei2023instructerc,hung2023beyond,devlin2018bert}, and facial expression recognition methods~\cite{jiang2020dfew,wang2020suppressing,ngwe2023patt} have shown their strengths on single modality emotion perception. 
However, beyond single modality, real-world emotional expressions often involve textual, visual, and auditory information, but these important cues of human emotions are overlooked in traditional methods. 

To utilize the information from different modalities, previous multimodal methods have focused on feature interaction and modality completion~\cite{li2023decoupled,wang2024incomplete,zadeh2018multimodal,zhang2023learning,zhou2019exploring,zhao2023prompting,sun2023mae,chumachenko2024mma}, but lack of reasoning abilities. 
Despite significant advancements in multimodal large language models (MLLMs)\cite{openai2023gpt4v,wang2024qwen2,li2024llava,chen2024expanding,Qwen2Audio}, which have demonstrated impressive performance in visual-language and audio-language analysis, challenges persist in accurately recognizing emotions. 

As shown in Figure~\ref{fig:sta}, these methods often fall short in precise emotional analysis even with step-by-step analyses. 
Recently, Emotion-LLaMA~\cite{cheng2024emotion}, AffectGPT~\cite{lian2024affectgpt}, and EmoLLM~\cite{lian2024affectgpt} tackled these issues by integrating universal vision encoders, an audio encoder, and an LLM decoder; however, they lack facial modeling and struggle to capture fine-grained facial information effectively. As a result, these limitations lead to sub-optimal performance in real-world scenarios.

We argue that another factor hindering the effectiveness of MLLMs is the absence of multimodal emotion instruction datasets with diverse data distributions and detailed annotations. 
The inconsistency in labeling standards among individuals and varying definitions of expressions also pose significant challenges. 
Different cultural contexts influence how emotions are expressed, which complicates the acquisition of emotion data and necessitates expert annotation.

To acquire high-quality multimodal emotion instruction tuning data, we curate two types of emotion datasets from existing labeled emotion datasets. For each video clip, we utilize a general video MLLM with strong performance to describe the general visual cues, including surroundings and characters. Next, we extract the face tracklets and use an age/gender estimator along with general video MLLM to capture facial attributes and fine-grained expressions, respectively. 
Then, we use GPT-3.5~\cite{openai2023chatgpt} to ensure consistency and discard descriptions that do not align with other clues, generating the final multimodal emotion reasoning and open-vocabulary labels. 
As shown in Figure~\ref{fig:intro}, the annotations generated by our processing pipeline are highly detailed. 
By aligning the detailed annotations with the coarse-grained ground-truth classification labels and eliminating the inconsistency, the quality of the annotations is significantly enhanced.
Finally, we have collected 24,137 self-reviewed videos with high alignment scores, naming this the self-reviewed emotion (SRE) dataset. Additionally, we evenly selected 3,500 videos for manual verification, resulting in a human-reviewed emotion (HRE) dataset.

With the high-quality detailed-annotated datasets, we adopt an effective open-source video MLLM as our base model and integrate the audio and face encoder to capture fine-grained auditory clues and facial movement for emotion analysis. Specifically, we first align the audio feature from whisper-large-v3~\cite{radford2023robust}, and the facial feature from FaceXFormer~\cite{narayan2024facexformer} to the general video MLLM embedding space. 
Subsequently, we leverage our high-quality data to simultaneously train the projectors of three encoders along with the LLM decoder. Our proposed model, named Omni-Emotion, shows strong performance in multiple tasks and is capable of processing information from various modalities.
Our main contributions are as follows.

\noindent\textbf{High-quality dataset construction.} 
We develop an effective and scalable approach to construct high-quality benchmark for emotion reason analysis and open-vocabulary emotion recognition based on existing emotion recognition datasets. The SRE dataset includes 24,137 videos filtered by GPT scoring, and the HRE dataset includes 3,500 videos that have been manually verified.

\noindent\textbf{Integrating video MLLM with auditory and facial information.} 
We propose to integrate additional face and audio encoders with the existing video MLLM model for better emotion analysis.  
We also build a three-stage training process to effectively unify audio encoder and fine-grained face encoder into Video MLLM.

\noindent\textbf{State-of-the-Art performance.} Our model achieves state-of-the-art (SOTA) results across various emotion analysis tasks. For in-the-wild emotion recognition, the unweighted average recall (UAR) on DFEW~\cite{jiang2020dfew} and MAFW~\cite{liu2022mafw} of our model are 68.80/53.81 respectively. In open-vocabulary emotion recognition task, our average score in EMER-OV~\cite{lian2023explainable} is 65.9. In the emotion reason analysis dataset EMER~\cite{lian2023explainable} , our clue overlap score is 8.22.  Our model outperforms existing methods by a clear margin.

\section{Related works}

\noindent \textbf{Multimodal Large Language Models (MLLMs).}
Recently, MLLMs~\cite{alayrac2022flamingo, qwen, ouyang2022instruct-tuning, shikra, vicuna2023, peng2023kosmos, wang2023visionllm,li2023blip} have garnered significant attention for their impressive inferential capabilities. 
Image-based MLLMs~\cite{chen2023minigpt, liu2024visual, zhu2023minigpt,tong2024cambrian} transformed image-text pairs into instruction-following data and study alignment problems between text and image features, demonstrating impressive multimodal capabilities. Despite these advancements, 
while most video MLLMs\cite{zhang2023video,li2024llava,li2024mvbench} utilize pretrained general vision encoders like CLIP \cite{radford2021learning}, SigLIP~\cite{zhai2023sigmoid}, and InternVideo2~\cite{wang2024internvideo2} for general domain applications, they encounter difficulties in understanding emotions from audio cues and subtle facial expressions.
The limitation is also due to the lack of specialized training on datasets that integrate multimodal emotional cues and comprehensive emotion-related knowledge.
Compared to these methods, our proposed method can extend strong video MLLM to audio modality with a strong audio encoder, and enable MLLM to capture fine-grained facial cues, thereby facilitating the effective analysis of human emotions in video.

\begin{figure*}[t!]
      \centering
    \includegraphics[height=6.3cm,trim=0 0 0 0,clip]{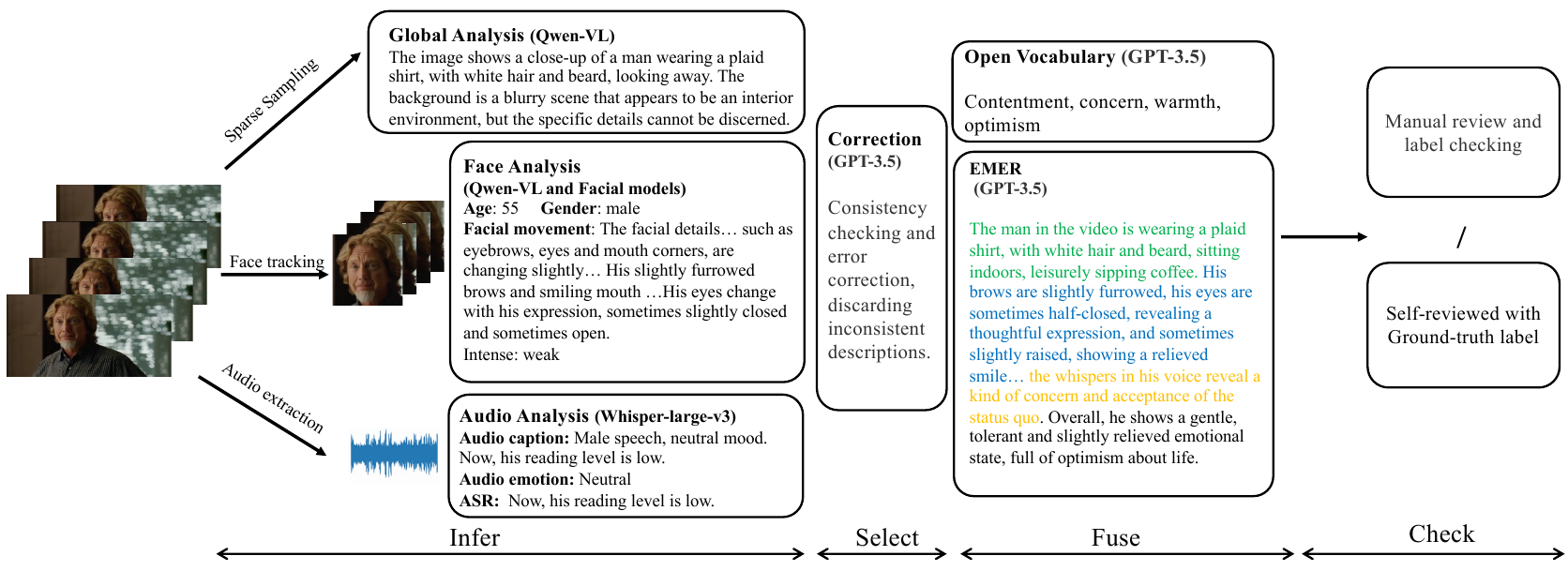}
      \caption{
      The processing pipeline of our proposed datasets consists of three aspects, including the extraction of global visual information, fine-grained facial details, and audio cues. 
      First, we conduct sparse sampling at 1 FPS for each video, using Qwen2-VL to capture global visual information.
      For detailed facial analysis, a face detector and tracking method are employed to extract face tracklets, followed by dense sampling and analysis using an age/gender estimator and Qwen2-VL.
      Lastly, we use Whisper-large-v3 to analyze the audio cues.
      We then employ GPT-3.5 to ensure consistency and accuracy, discarding descriptions that do not align with other clues. 
      Based on these results, we extract explainable reasons, open vocabulary descriptions, and intensity for emotion of each video. 
      Finally, we utilize GPT-3.5 to review the generated data referring to existing coarse-grained classification labels and obtain the self-reviewed emotion (SRE) dataset.
      We select partial data from SRE dataset and perform manual verification to build the human-reviewed emotion (HRE) dataset.
      }
      \label{fig:datacot}
\end{figure*}

\noindent \textbf{Emotion analysis.}  
Emotion and facial expression recognition have predominantly depended on datasets collected under controlled conditions, often leading to suboptimal performance in real-world applications. Some large dynamic facial expressions in-the-wild datasets~\cite{jiang2020dfew,liu2022mafw,wang2022ferv39k} facilitate the availability. 
While the previous methods~\cite{zhao2023prompting,sun2023mae,chumachenko2024mma} focus on emotion recognition, human emotions are complex and often involve multiple emotions in various situations. Additionally, understanding the underlying causes of these emotions is also crucial.
EmoVIT \cite{xie2024emovit} generated visual emotion instruction data for training; however, it lacks audio information. On the other hand, EMRE \cite{lian2023explainable} offered detailed annotations with explainable reasons for emotions but included only 332 samples. AffectGPT \cite{lian2023explainable} and Emotion-LLaMA \cite{cheng2024emotion} collected a large volume of inference annotations, yet these annotations lack quality assurance. Compared to Emotion-LLaMA, our collected data with self-review by GPT-3.5 using ground-truth labels and manual reviews. Additionally, we extend existing video MLLMs to capture detailed facial and auditory information.

\section{Self-Reviewed Emotion Dataset}
\label{sec:dataset}

In this section, we introduce the processing of emotional data and the curation of our proposed datasets.
Existing emotion datasets~\cite{jiang2020dfew,liu2022mafw,wang2022ferv39k} predominantly use categorical labels, which often fail to capture the nuanced emotions and expressions in videos. Although initiatives like EMER~\cite{lian2023explainable} attempt to use open vocabulary and detailed reasons for annotation, their scale is limited to only 332 entries because the process of emotion annotation demands highly skilled emotion knowledge, making it prohibitively expensive. Emotion-LLaMA~\cite{cheng2024emotion} introduces the MERR dataset with model-inferred descriptions but lacks thorough quality assessments. Besides, their explanations for the emotions are not detailed enough to fully capture emotional clues.

To collect a substantial and reliable dataset for accurately analyzing emotions in videos, 
we propose to leverage existing MLLM and advanced facial analysis model to provide detailed, comprehensive annotations to enrich existing emotion datasets, and perform self-reviewing by comparing the ground-truth classification labels and the inferring detailed annotations.
Our approach entails reasoning based on datasets including the training set from DFEW~\cite{jiang2020dfew}, MAFW~\cite{liu2022mafw}, MER24~\cite{lian2024mer}, CAER~\cite{lee2019context}, AFEW-VA~\cite{kossaifi2017afew}, RAVDESS~\cite{livingstone2018ryerson}, and FERV39K\cite{wang2022ferv39k}, exploring global information, detailed facial features, and audio cues to extract emotional clues.

Our method involves three aspects, as illustrated in Figure~\ref{fig:datacot}. First, we perform low-density sampling at 1 FPS for each video and use Qwen2-VL-72B for video description. Simultaneously, we use face detection~\cite{deng2019retinaface} and tracking~\cite{wojke2017simple} methods to extract each facial tracklet and conduct fine-grained analysis of age, gender, and facial expressions.
Specifically, MiVOLO~\cite{kuprashevich2023mivolo} is employed for age and gender estimation, and Qwen2-VL-72B~\cite{wang2024qwen2} is employed to describe the detailed facial movement. Finally, we extract audio tracks and utilize Whisper-large-v3~\cite{radford2023robust} for captioning, automatic speech recognition (ASR), and emotion recognition. The consistency of the information obtained from these steps is verified using GPT-3.5~\cite{openai2023chatgpt}. Inconsistent or uncertain data is discarded, ensuring high-quality intermediate results.

\subsection{Self-Reviewed Emotion Dataset Curation}

\begin{figure}[t]
      \centering
      \includegraphics[height=5cm,trim=0 0 0 0,clip]{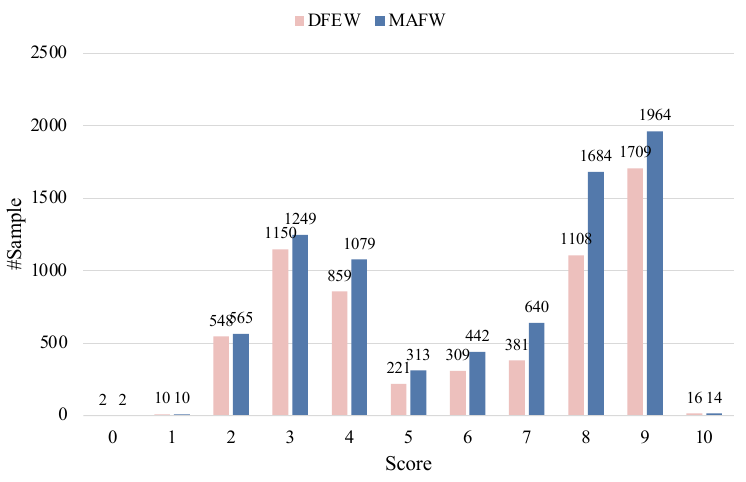}
      \caption{Statistics of the GPT-based scores assessing the alignment between emotion reasoning descriptions and ground-truth classification labels on the MAFW~\cite{liu2022mafw} and DFEW~\cite{jiang2020dfew} datasets. The scores range from 0 to 10, with lower scores indicating greater deviation from the ground truth. Samples scoring below 5 are considered misaligned.
      }
      \label{fig:sta}
\end{figure}

Building on these results, we derive explainable reasoning clues for emotions by analyzing them from visual, auditory, and textual perspectives, and obtain open-vocabulary annotations.
To ensure quality, we utilize the original labels from these datasets and prompt GPT-3.5 to assess the alignment degree between the ground-truth labels and explainable reason statements on a 0-10 scoring scale. In Figure~\ref{fig:sta}, we visualize the score statistics on DFEW~\cite{jiang2020dfew} and MAFW~\cite{liu2022mafw}. Despite rigorous processing, achieving high agreement scores remains challenging, underscoring the complexity of emotion recognition. Accordingly, we filter out samples with a score below 5.

\begin{figure}[t]
      \centering
      \includegraphics[height=5cm,trim=0 0 0 0,clip]{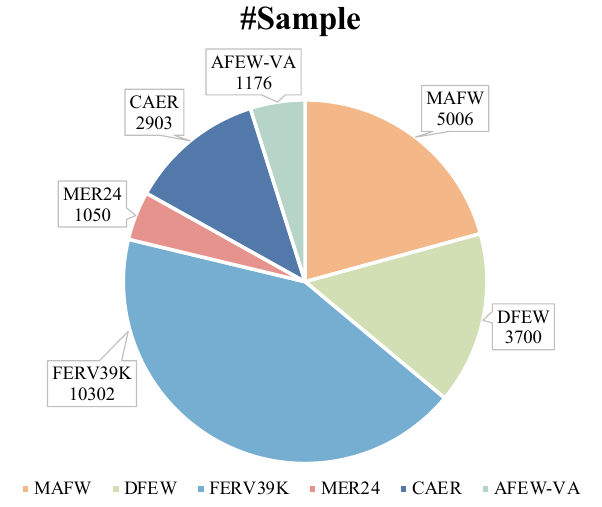}
      \caption{The distribution of sample sizes across several emotion recognition datasets. FERV39K has the largest portion with 10,302 samples, followed by MAFW with 5,006. DFEW and CAER contribute 3,700 and 2,903 samples, respectively. Meanwhile, MER24 and AFEW-VA add 1,050 and 1,176 samples, respectively, providing a diverse distribution for emotion analysis.
      }
      \label{fig:datacount}
\end{figure}

After filtering with a score threshold, our self-reviewed annotation process extends the traditional emotion classification to emotion analysis from multiple perspectives while maintaining a high level of agreement.
We further remove the RAVDESS dataset considering the video clips are collected from the indoor studio.
Finally, the self-reviewed dataset comprises 24,137 samples, as shown in Figure~\ref{fig:datacount}, and we refer to this dataset as the \textbf{Self-Reviewed Emotion (SRE) dataset}.
\subsection{Human-Reviewed Emotion Dataset Curation}
In addition, we have constructed a human-annotated dataset by randomly selecting 700 samples from each training set of these datasets, ensuring no overlap with the SRE dataset, namely \textbf{human-reviewed emotion (HRE) dataset}. Ultimately, we assembled a higher-quality training dataset consisting of 3500 samples. Our SRE and HRE dataset includes a wide range of samples from diverse sources.


\begin{figure*}[t]
      \centering
      \includegraphics[height=7cm,trim=0 0 0 0,clip]{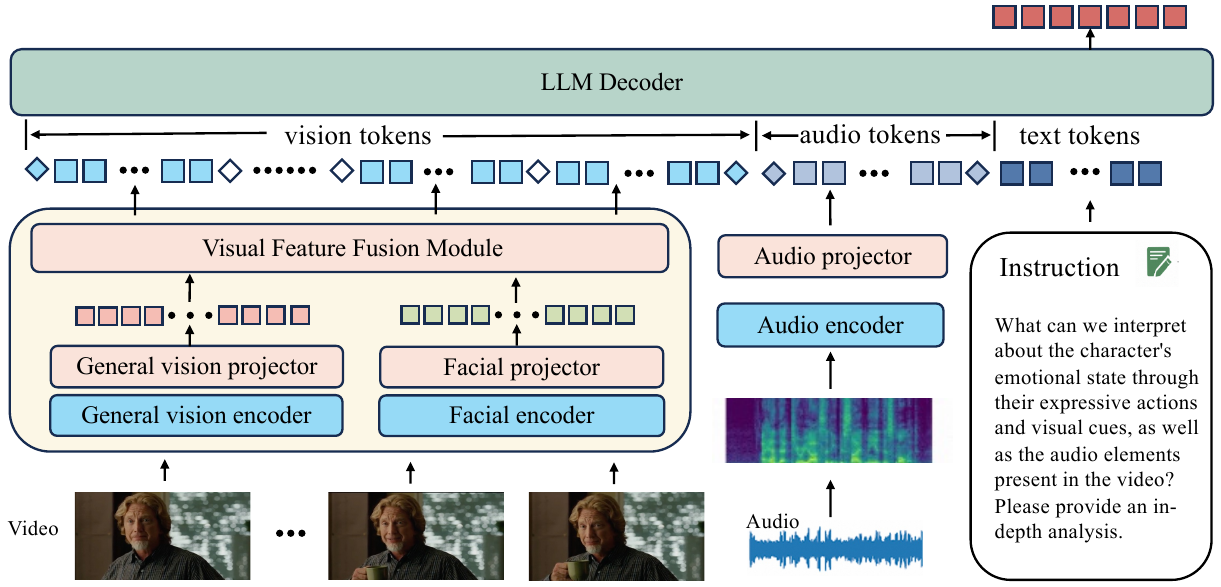}
      \caption{Illustration of our proposed Omni-Emotion MLLM. Our model includes an LLM decoder, a general vision encoder for extracting visual clues, a facial encoder for capturing fine-grained facial information, and an audio encoder for processing auditory input. 
      Our Omni-Emotion MLLM can leverage facial and audio details to better comprehend human emotion.
      }
      \label{fig:model}
\end{figure*}

\section{Omni-Emotion MLLM}
While existing open-source video MLLMs have achieved significant improvements in video perception tasks, they fall short in emotion analysis due to their limited ability to process audio inputs and capture fine-grained facial information.
To mitigate these limitations, we propose to extend the video MLLM to audio modality first and then introduce the facial detailed modeling ability to capture fine-grained facial clues.
An illustration of our proposed Omni-Emotion MLLM is in Figure~\ref{fig:model}.

\subsection{Extending Video MLLM to Audio Modality}

We use the audio encoder from Whisper-large-v3~\cite{radford2023robust} as our audio encoder, which is a state-of-the-art model for automatic speech recognition (ASR) and speech translation. 
Given the paired data $(\mathbf{a}, \mathbf{x})$, where the $\mathbf{a}$ and $\mathbf{x}$ denote the audio sequences and text sequences,  the training objective focuses on maximizing the probability of the next text token, formulated as:
\begin{equation}
\mathcal{P}_\theta(x_t|\mathbf{x}_{<t}, \text{Encoder}_{\phi}(\mathbf{a})),
\end{equation}
where $\theta$ and $\phi$ denote the parameters of the LLM decoder and audio encoder, respectively.
The probability is conditioned on the audio representations and preceding text sequence $x<t$, aiming to bridge audio features with language features.

Specifically, the initialization of the audio encoder is loaded from the encoder weights of the Whisper-large-v3 model. We resample each audio data to a frequency of 16KHz and convert the raw waveform into 128 channel mel-spectrogram using a window size of 25ms and a hop size of 10ms. We use a pooling layer with a stride 3 to reduce to token length, and each token approximately corresponds to a 60ms segment of the original audio signal. Each audio feature is fed into the audio projector, which consists of two linear layers. The output dimension of the projector is designed to match the hidden embedding size of LLM.

\subsection{Detail Facial Modeling}
Most video MLLMs utilize general visual encoders, such as SigLip~\cite{zhai2023sigmoid}, CLIP~\cite{radford2021learning}, or InternVideo2~\cite{wang2024internvideo2}, to extract general visual features from videos. However, these encoders do not specifically focus on extracting person-related or face-related features, and the features they capture often contain many other objects or background cues. While these cues are valuable for general tasks, they are not particularly important for analyzing human emotions.

One of the crucial clues for analyzing emotions is facial information. 
Hence, we introduce an additional facial feature encoder in our Omni-Emotion MLLM and adopt the encoder in FaceXFormer~\cite{narayan2024facexformer} as our facial feature encoder. FaceXFormer is an end-to-end encoder-decoder transformer architecture designed for a wide range of facial analysis tasks, including face parsing, landmark detection, head pose estimation, and attribute recognition. 
We believe its encoder could provide robust and generalized face representation capable of handling images in the wild. 

For an input video frame, we first extract the fine-grained multi-scale features; and then a lightweight MLP-fusion module generates a fused face representation from the multi-scale features. The transformed features are finally concatenated together, flattened, and fed into the facial projector, which consists of two linear layers. The output dimension of the projector is equal to the hidden embedding size of LLM.

\subsection{Visual Feature Fusion Module}
\begin{figure}[t]
      \centering
      \includegraphics[height=5.5cm,trim=0 0 0 0,clip]{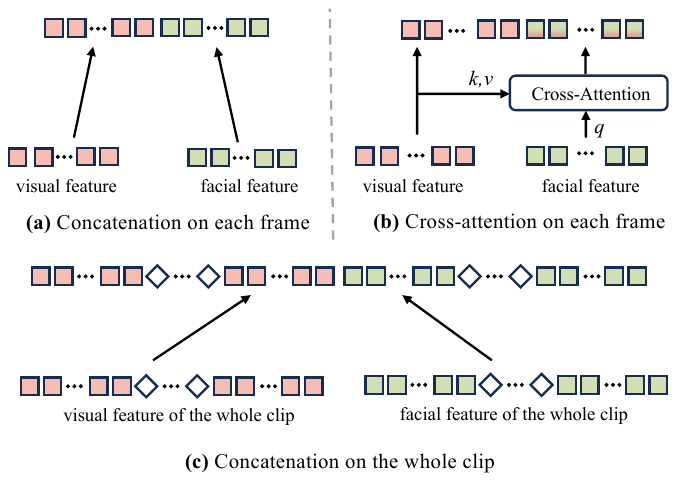}
      \caption{
      Illustration of the visual feature fusion methods: (a) The facial feature are concatenated with the general visual feature for each frame. (b) Merge facial and general visual features per frame in a cross-attention layer, while passing the general visual feature to the LLM decoder to retain general knowledge. (c) Concatenate the facial feature of the entire video clip with its corresponding general visual feature.
      }
      \label{fig:fusion}
\end{figure}

To maintain general knowledge while capturing fine-grained facial details, we develop a vision fusion module to integrate the facial features with general visual features. 
Given the different lengths of these features, approaches like the weighted sum or concatenation along the channel dimension are not applicable.

As shown in Figure~\ref{fig:fusion}, we examine three methods: concatenation along sequence length either at the frame level or video level, and employing a cross-attention layer.
Figure~\ref{fig:fusion} (a) shows the first approach that directly concatenates facial and visual features for each frame, ensuring that each frame contains both visual and facial features. The modeling of the relation between facial and visual information is left to the LLM decoder.
The second approach uses cross-attention, with facial features act as the query and visual features as the keys and values, to extract person-related information from the visual features, as shown in Figure~\ref{fig:fusion} (b). The third approach is to separately extracts the visual and facial features for the entire video and then concatenating them. This fusion method has the least effect on the original video MLLM.

The fusion features serve as our final visual tokens, which are then combined with embeddings from other modalities and fed into the LLM decoder.

\subsection{Training Detail}
\label{sec:training}
Our proposed Omni-emotion MLLM is built on LLaVA-OneVision-7B~\cite{li2024llava}.
We keep the general visual encoder and projector from LLaVA-OneVision-7B and introduce an additional facial and audio encoder with the corresponding projectors.
To effectively unify the information from three encoders, we first align the audio and face features to the original MLLM feature space.

\noindent \textbf{Phase 1: Audio Feature Alignment.} 
To align the audio features with the text embeddings, we use CommonVoice-15~\cite{ardila2019common}, WavCaps~\cite{mei2024wavcaps}, and VGGSound~\cite{chen2020vggsound} as audio-text alignment datasets. We format the original labels into diverse instruction tuning formats instead of task tagging as in~\cite{radford2023robust}.
\begin{tcolorbox}[colback=gray!20, colframe=black!70, arc=2mm, auto outer arc, boxrule=0.3mm, width=0.47\textwidth, title=Example of audio QA pair]
    \raggedright 
    \textit{\textbf{Question}:Listen to this audio clip and provide its caption in English.}\\ 
    \textit{ }\\
    \textit{\textbf{Answer}: Wind and a man speaking are heard, accompanied by buzzing and ticking.}
\end{tcolorbox}
\noindent During this training phase, we freeze the rest of the model and only train the audio projector for one epoch, using a batch size of 256 and a learning rate of 1e-3. As a result, the audio features are aligned with the text embedding space.

\noindent \textbf{Phase 2: Facial Feature Alignment.} 
In the second phase, we reformat the origin emotion recognition data into the instruction-tuning format. We use the training set from seven emotion classification datasets and ask MLLM to choose the most matching option from the option list. 
The classification annotation is formatted as follows:
\begin{tcolorbox}[colback=gray!20, colframe=black!70, arc=2mm, auto outer arc, boxrule=0.3mm, width=0.47\textwidth, title=Example of emotion recognition  QA pair]
    \raggedright 
    \textit{\textbf{Question}: As an emotional recognition expert, when you observe the video, what is the primary emotion exhibited by the characters? \ul{surprise, neutral, fear, angry, happy, sad, disgust}}.\\ 
    \textit{ }\\
    \textit{\textbf{Answer}: neutral}.
\end{tcolorbox}
\noindent In this training phase, we only use the general visual and facial features, freeze the rest of the model, and only train the facial projector for 1 epoch.  The learning rate is set to 1e-3, and the total batch size is 256.

\noindent \textbf{Phase 3: Multimodal Instruction Tuning.} 
In this training phase, we leverage the general visual and facial features of each frame. For the features of each frame, we add a learnable parameter to help the model distinguish between different frame sequences, thereby improving temporal modeling. Additionally, for both visual and audio features, we use specific tokens to label them. The multimodal template is denoted as:
\begin{tcolorbox}[colback=gray!20, colframe=black!70, arc=2mm, auto outer arc, boxrule=0.5mm, width=0.47\textwidth, title=Example of modality special token]
    \raggedright 
    \textit{$<$vi\_start$>$$<$vision\_feature$>$$<$vi\_end$>$$<$au\_start$>$\\$<$audio\_feature$>$$<$au\_end$>$}\\ 

\end{tcolorbox}

\noindent The wrapped multimodal embedding is further wrapped by the original chat template of the LLM. If one modality is absent, we use a zero tensor to replace it.
This approach helps the model differentiate between features from different modalities, preventing information confusion in cases where one modality may be absent (\ie, preventing the model from being uncertain whether an embedding originates from the visual, auditory, or textual domain).

In this phase, \yx{our goal is to develop an omnimodal emotion model for \textbf{open-vocabulary emotion recognition} and \textbf{multimodal emotion reasoning} using the proposed SRE and HRE datasets.} We fine-tune three projectors and the LLM decoder for a total of 3 epochs. We sample 8 frames for each video and use the concatenation of the general visual features and facial features at the video level as the default fusion feature. The learning rate and batch size are 1e-5 and 128, respectively.

\section{Experiments}
We conduct experiments across distinct settings to evaluate the effectiveness of our proposed method and dataset. 
Since human emotions are often mixed or extend beyond these predefined categories, (1) we follow AffectGPT~\cite{lian2024affectgpt} and employ all the open-vocabulary labeled data provided in the EMER dataset to assess open-vocabulary emotion recognition capabilities, which include 332 samples in total. We refer to this setting as ``EMER-OV'' in the following text. (2) Additionally, to further analyze the underlying reasons for emotions, we leverage the explainable multimodal emotion recognition descriptions in EMER to evaluate the model’s emotion reasoning abilities.
\yx{(3) Finally, we report additional results in emotion recognition task.}

\subsection{Open-Vocabulary Emotion Recognition} 
Following AffectGPT~\cite{lian2024affectgpt}, we use the open-vocabulary labels in EMER-OV~\cite{lian2023explainable} to evaluate the open-set prediction performance of our model. The EMER-OV includes 332 open-vocabulary-labeled samples. In this experiment, we compare our method with AffectGPT~\cite{lian2024affectgpt}. 

We use the same method as~\cite{lian2024affectgpt} to group all labels and convert the emotion grouping information into a function that can map each label to its group ID. And then map each label into its group ID. Finally, we can compute the recall, precision, and average score between the ground-truth group $\mathcal{Y}$ and the predicted group $\hat{\mathcal{Y}}$ as follows:
\begin{equation}
\mbox{Precision}_{\mbox{s}} = \frac{|\mathcal{Y} \cap \hat{\mathcal{Y}}|}{|\hat{\mathcal{Y}}|}, \;\mbox{Recall}_{\mbox{s}} = \frac{|\mathcal{Y} \cap \hat{\mathcal{Y}}|}{|\mathcal{Y}|}.
\end{equation}
For more computation details, please refer to~\cite{lian2024affectgpt,lian2023explainable}. 

As shown in Table~\ref{table:ov}, our model significantly outperforms AffectGPT, especially in terms of recall. Besides, our model can return a richer set of emotional descriptors, such as ``Excited'', ``Calm'', ``Frustration'', and others. 
This experiment confirms the quality and diversity of our collected datasets and highlights the potential and effectiveness of our method as a promising solution for real-world applications.

\begin{table}[t]
    \resizebox{0.95\textwidth}{!}{%

    \centering
    \begin{tabular}{l|c|c|c}
    \hline
    Method & Avg & Accuracy (\%) & Recall (\%) \\ \hline

    AffectGPT~\cite{lian2024affectgpt} & 56.6 & 47.5  & 65.7 \\  \hline
    Omni-Emotion(v) & 53.4 & 48.8  & 58.0 \\
    Omni-Emotion(v+a) & 63.1 & 61.7  & 64.4 \\ 
    \textbf{Omni-Emotion(v+a+f)} & \textbf{65.9} & \textbf{61.3}  & \textbf{70.6} \\ \hline
    \end{tabular}
    \caption{Results on open vocabulary setting of EMER-OV~\cite{lian2023explainable} dataset. ``v'', ``a'', and ``f'' represent the general visual, auditory, and facial features, respectively.}
     \label{table:ov}
    }
\end{table}

\subsection{Multimodal Emotion Reasoning} 
The explainable reason description in EMER dataset is significantly different from traditional emotion recognition datasets due to the inclusion of additional multimodal emotion clues, such as facial micro-expressions, speech tone, and contextual video information, alongside the emotion categories. To evaluate the emotional reasoning abilities of various MLLMs on the EMER dataset, we use the same setting as the Emotion-LLaMA~\cite{cheng2024emotion} and use ChatGPT to score their predictions, focusing on two metrics: (1) the overlap between the emotion-related cues, and (2) the overlap in the summarized emotional states. 
As shown in Table~\ref{table:emer}, our proposed method achieves the highest on both metrics. These experiments validate the effectiveness of our proposed method and the quality of SRE and HRE datasets.

\begin{table}[t]

\begin{tabular}{lcc}
\toprule
Models & Clue Overlap & Label Overlap\\
\midrule
VideoChat-Text ~\cite{li2023videochat}      & {6.42}    & {3.94}\\
Video-LLaMA ~\cite{zhang2023video}         & {6.64}    & {4.89}\\
Video-ChatGPT ~\cite{maaz2023video}       & {6.95}    & {5.74}\\
PandaGPT ~\cite{su2023pandagpt}            & {7.14}    & {5.51}\\
VideoChat-Embed ~\cite{li2023videochat}     & {7.15}    & {5.65}\\
Valley ~\cite{luo2023valley}              & {7.24}    & {5.77}\\

Emotion-LLaMA~\cite{cheng2024emotion} & 7.83    & 6.25\\
\textbf{Omni-Emotion (Ours)} & \textbf{8.22}    & \textbf{6.78}\\
\bottomrule
\end{tabular}
\caption{Comparison of multimodal emotion reasoning results on EMER.}
 \vspace{-8pt}
\label{table:emer}

\end{table}

\begin{table}[t]
    \resizebox{0.95\textwidth}{!}{%

    \centering
    \begin{tabular}{l|c|c|c|c}
    \hline
     Dataset & \multicolumn{3}{c|}{EMER-OV (\%)} & EMER\\ \hline
    Method & Avg. & Acc. & Recall  & Clue\\ \hline
    Concatenation on per frame & 63.4 & 60.1  & 66.6 & 7.93\\
    Cross-attention on per frame & 60.7 & 58.1  & 63.2 & 7.31\\ 
    Concatenation on video level & 65.9 & 61.3  & 70.6 & 8.22\\ \hline
    
    \end{tabular}
    \caption{Results   on open vocabulary setting of EMER~\cite{lian2023explainable} dataset.}
        \label{table:fusion}
    }
\end{table}

\begin{table*}[t]
\caption{An example of multimodal emotion reasoning comparing Omni-Emotion with other MLLMs. We highlight the incorrect reasoning with red color.}
\centering  
\scalebox{0.85}{
    \begin{tabular}{l|p{16cm} }
\toprule
\multicolumn{2}{l}{ The reasoning results of ``sample\_00000342.mp4'' In EMER~\cite{lian2023explainable} dataset} \\
\midrule
& {\includegraphics[height=1.9cm]{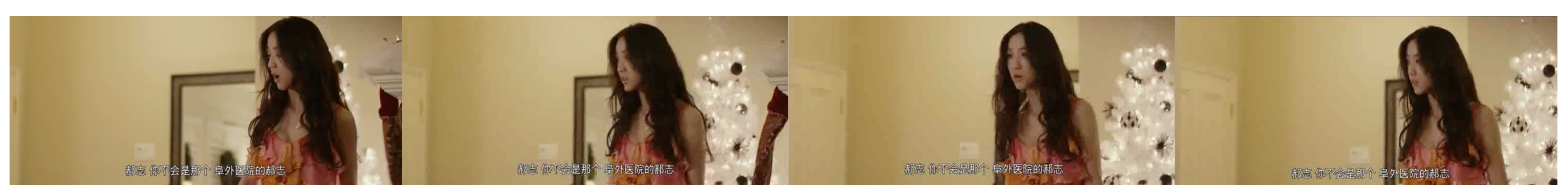}} \\
Subtitle (En) & {Hao Zhi, are you the Hao Zhi from Fuwai Hospital?} \\
\midrule
Ground Truth & {In the video, it is impossible to determine the emotional state of the person. In the audio, when expressing ``Hao Zhi'', the tone rises and the tone sounds surprised. In the text, the subtitle says, ``Hao Zhi, are you the Hao Zhi from Fuwai Hospital?'' Based on the description of the rising tone and the surprised tone in the audio clue, as well as the inability to determine the emotional state of the person in the video clue, we can infer that this sentence may carry a sense of surprise or suspicion. The woman may be surprised or uncertain about Hao Zhi's identity, hence expressing this question.}  \\
\midrule
Emotion-LLaMA & {in the video, the young woman's facial expression and body language suggest a sense of surprise, \textcolor[rgb]{1,0,0}{as she is standing in front of a mirror, looking at herself, possibly admiring her appearance or reflecting on her thoughts her eyebrows are slightly raised}, which can indicate a moment of curiosity or interest her mouth is open, which may be a result of her surprise or simply an expression of concentration her posture appears relaxed, with her arms at her sides, and her hands not visible, \textcolor[rgb]{1,0,0}{indicating a lack of tension or anxiety her overall demeanor suggests a sense of calmness and introspection, as she is engaging in self-reflection or admiring her own appearance, which is often associated with a feeling of surprise}}  \\
\midrule
Omni-Emotion & {In the video, a young woman dressed in a pink dress stands in a cozy indoor setting. Her expression shifts from surprise to confusion, ultimately revealing a look of concern. Her brows are slightly furrowed, her eyes wide open, and her lips slightly parted, as if she is trying to comprehend or respond to some unexpected information. This emotional transition suggests that she may have received shocking news or encountered an unforeseen situation, leaving her feeling puzzled and worried. The entire process showcases her immediate reaction to external stimuli and her inner turmoil.}  \\
\midrule

\end{tabular}
}
\label{table:case}
\end{table*}

\subsection{Analysis}

\noindent\textbf{Ablation study on each modality.}
In Table~\ref{table:ov}, we perform modality analysis in EMER-OV. We can observe that the auditory modality significantly improves recall and accuracy. Furthermore, the combination of ``vision+audio+face'' achieves the best average performance, while the accuracy is slightly lower than the ``vision+audio'' combination. 
These experimental results reveal that the fine-grained face clue is also important for multimodal open-vocabulary emotion recognition.

\noindent\textbf{Analysis on visual feature fusion.}
We analyze the visual fusion modules on EMER-OV and EMER datasets, and present the results in Table~\ref{table:fusion}. While the concatenation on the video level achieves the highest score in both settings, we believe that this design is most compatible with the original video MLLM. However, given the significant gap in our training data compared to the general video MLLM, we will further explore frame fusion methods.

\begin{table}[t]
    \resizebox{\textwidth}{!}{%
    \centering
    \begin{tabular}{l|ccccc}
    \hline
    Dataset & \multicolumn{2}{c}{DFEW~\cite{jiang2020dfew}}  & \multicolumn{2}{c}{MAFW~\cite{liu2022mafw}} & \\ \hline
    Methods& UAR & WAR &  UAR & WAR & M \\  \hline
    Wav2Vec2.0 \cite{baevski2020wav2vec} & 36.15&43.05&21.59&29.69&A\\
HuBERT \cite{hsu2021hubert}&35.98&43.24&25.00&32.60&A\\

DFER-CLIP  \cite{zhao2023prompting}     & 59.61  & 71.25       & 38.89   & 52.55 & V   \\
SVFAP \cite{sun2024svfap} & 62.83 & 74.27 & 41.19 & 54.28 & V  \\

    MAE-DFER~\cite{sun2023mae} & 63.41 & 74.43 & 41.62 & 54.31 &V\\
        Finecliper~\cite{chen2024finecliper} & 65.98 & 76.21 & 45.01 & 56.91 & V\\
    TMEP \cite{zhang2023transformer} & 57.16 & 68.85 &37.17 & 51.15 & AV \\
    HiCMAE   \cite{sun2024hicmae}       &   63.76    &  75.01   & 42.65   &  56.17   & AV   \\ 
    Emotion-LLaMA~\cite{cheng2024emotion} & 64.21 & 77.06 & - & - & AV \\
    MMA-DFER~\cite{chumachenko2024mma} & 66.01 & 77.51 & 44.11 & 58.52 & AV\\ \hline

    Omni-Emotion (Ours) & 68.80 & 78.35 & 53.81 & 64.23 & AV \\ \hline

    \end{tabular}
    \caption{Comparison to SOTA methods in two large in-the-wild emotion datasets (DFEW and MAFW). M denotes the modality}

    \vspace{-8pt}
    \label{table:emotionr}

    }
\end{table}

\noindent\textbf{Qualitative result}
We present an example of qualitative comparison with Emotion-LLaMA in Table~\ref{table:case}. Our model provides more accurate descriptions of surroundings, the character,  and a more detailed understanding of facial expressions, enabling precise inference of the reasons behind the emotions. More examples are provided in the supplementary materials.

\subsection{Emotion Recognition Evaluation.} 
We utilized two in-the-wild datasets DFEW~\cite{jiang2020dfew} and MAFW~\cite{liu2022mafw} containing audio information to evaluate the emotion recognition performance of our model. The test sets for DFEW and MAFW contain 7 and 11 emotion categories, respectively. 
In this task, we reuse question-answer pairs data as discussed in \textit{phase 2} to perform instruction tuning. We then applied instruction tuning for projectors and the LLM decoder using this training data for 3 epochs. 
We compared our model with other MLLMs and state-of-the-art (SOTA) methods using unweighted average recall (UAR) and weighted average recall (WAR) as metrics. 
The experimental results are shown in Table~\ref{table:emotionr}. 
Compared to single-modality approaches, multimodal methods show a significant advantage. Our proposed method achieves state-of-the-art performance on both datasets, with a notable margin on the MAFW dataset. The MAFW dataset includes 11 emotion classes, presenting a greater challenge that requires capturing fine-grained facial expressions and audio cues.

\subsection{Limitations}
Our SRE and HRE datasets are curated from existing datasets, yet the video diversity remains far from that encountered in real-world applications. For instance, in most scenarios, users tend to capture data using the front-facing camera, leading to a different data distribution with existing sources.
Moreover, despite our proposed method attaining SOTA performance, its accuracy in handling mixed emotions still falls short. Furthermore, the size of our model and the input token length also pose substantial challenges for practical deployment.

\section{Conclusion}
Current MLLMs also struggle with integrating audio and identifying facial micro-expressions due to insufficient detailed training data. To address this, we introduce a self-reviewed dataset with 24,137 samples and 3,500 manually annotated with detailed emotion annotations, enhancing model learning and generalization. We propose a novel method that incorporates audio and facial encoders into existing Video MLLMs, effectively capturing facial and audio cues. By aligning features in a unified space and using instruction tuning, our method, Omni-Emotion, achieves SOTA results in emotion recognition and reasoning.


\section{More Qualitative Results Details of  Datasets}
In Table~\ref{table:case1}, ~\ref{table:case3}, and ~\ref{table:case4}, we provide more visualization comparisons. Our model not only captures overall visual information and subtle facial expressions but also analyzes audio within the video. Our scores are significantly higher than those of Emotion-LLaMA~\cite{cheng2024emotion}, closely aligning with the details in the ground truth.

\section{Instructions Details of Training}
\noindent \textbf{Audio feature alignment.} In our audio training, we used a total of 1.5M audio instructions tuning data. Below, we present more instruction examples.

\begin{tcolorbox}[colback=gray!20, colframe=black!70, arc=2mm, auto outer arc, boxrule=0.3mm, width=0.47\textwidth, title= Examples of instructions for audio alignment.]
    \begin{itemize}[leftmargin=1mm]
\setlength{\itemsep}{1pt}
    \item \textit{Listen to this audio clip and provide its caption in English.}
    \item \textit{Could you summarise what's happening in this audio?}
    \item \textit{Please describe the audio in English.}
    \item \textit{Describe the following audio in a caption.}\\
    \item \textit{Write down the content of the speech you heard.}
    \item \textit{Give me the transcription of the speech you heard.}
    \item \textit{lease transcribe the speech into a written format.}
    \item \textit{Recognize the speech and write it down in a written format.}
    \item \textit{Recognize the speech and give me the transcription.}
    \end{itemize}
\end{tcolorbox}

\noindent \textbf{Facial feature alignment.} 
In our facial alignment training, we utilized about 130K data from 7 datasets (including the training sets from DFEW~\cite{jiang2020dfew}, MAFW~\cite{liu2022mafw}, MER24~\cite{lian2024mer}, CAER~\cite{lee2019context}, AFEW-VA~\cite{kossaifi2017afew}, RAVDESS~\cite{livingstone2018ryerson}, and FERV39K\cite{wang2022ferv39k}) for alignment training of the facial projector. Here are examples of instructions for facial alignment:

\begin{tcolorbox}[colback=gray!20, colframe=black!70, arc=2mm, auto outer arc, boxrule=0.3mm, width=0.47\textwidth, title= Examples of instructions for facial alignment.]
    \begin{itemize}[leftmargin=1mm]
\setlength{\itemsep}{1pt}
    \item \textit{Which emotion exhibited by the characters can you confirm as the primary emotion? \ul{disgust, angry, sad, surprise, neutral, fear, happy}}
    \item \textit{What primary emotion conveyed by the characters can you clearly identify? \ul{disgust, anxious, sad, surprise, neutral, fear, angry}}
    \item \textit{Which emotion can you recognize being expressed by the characters? \ul{disgust, sad, surprise, neutral, happy, angry}}
    \item \textit{What primary emotion conveyed by the characters can you clearly identify? \ul{fear, helpless, contemptuous, disappointed, happy, anxious, angry, sad}}
   
    \end{itemize}
\end{tcolorbox}

\noindent \textbf{Multimodal instruction tuning.} 
We used HRE and SRE datasets for multimodal instruction training, generating a total of about 100K instruction training data. Some of these instructions are shown below.
\begin{tcolorbox}[colback=gray!20, colframe=black!70, arc=2mm, auto outer arc, boxrule=0.3mm, width=0.47\textwidth, title= Examples of instructions for multimodal tuning.]
    \begin{itemize}[leftmargin=1mm]
\setlength{\itemsep}{1pt}
    \item \textit{What insights can we gain about the character's emotional state from their actions and facial expressions, as well as the accompanying audio and visual cues in the video? Please provide a detailed analysis.}
    \item \textit{Based on the character's physical actions and emotional expressions, along with the video's sound and visual context, what can we deduce about their emotional state? Please elaborate thoroughly.}
    \item \textit{What can we interpret about the character's emotional state through their expressive actions and visual cues, as well as the audio elements present in the video? Please provide an in-depth analysis.}
   
    \end{itemize}
\end{tcolorbox}

\section{Annotation comparison with other datasets}


In Table~\ref{table:diff}, we compare our dataset with EMER~\cite{lian2023explainable} and MERR~\cite{cheng2024emotion}. Our annotations are more aligned with EMER and more detailed compared to MERR. Additionally, we ensure the quality of our dataset through self-review against the original ground-truth labels and human review, which also allows for a larger scale of annotations. Additionally, we would like to correct the citation regarding the source of the EMER~\cite{lian2023explainable} data in Section 3 of the main text; the correct source is MER24~\cite{lian2024mer}. We apologize for this mistake.


\begin{table*}[t]
\caption{Examples of multimodal emotion reasoning comparing Omni-Emotion with other emotion MLLMs. We highlight the incorrect reasoning with red color.}
\centering  
\scalebox{0.85}{
    \begin{tabular}{l|p{16cm} }
\toprule
\multicolumn{2}{l}{ The reasoning results of ``sample\_00000342.mp4'' In EMER~\cite{lian2023explainable} dataset} \\
\midrule
& {\includegraphics[height=1.9cm]{figs/case.pdf}} \\
Subtitle (En) & {Hao Zhi, are you the Hao Zhi from Fuwai Hospital?} \\
\midrule
Ground Truth & {In the video, it is impossible to determine the emotional state of the person. In the audio, when expressing ``Hao Zhi'', the tone rises and the tone sounds surprised. In the text, the subtitle says, ``Hao Zhi, are you the Hao Zhi from Fuwai Hospital?'' Based on the description of the rising tone and the surprised tone in the audio clue, as well as the inability to determine the emotional state of the person in the video clue, we can infer that this sentence may carry a sense of surprise or suspicion. The woman may be surprised or uncertain about Hao Zhi's identity, hence expressing this question.}  \\
\midrule
Emotion-LLaMA & {
\textbf{Output:} in the video, the young woman's facial expression and body language suggest a sense of surprise, \textcolor[rgb]{1,0,0}{as she is standing in front of a mirror, looking at herself, possibly admiring her appearance or reflecting on her thoughts her eyebrows are slightly raised}, which can indicate a moment of curiosity or interest her mouth is open, which may be a result of her surprise or simply an expression of concentration her posture appears relaxed, with her arms at her sides, and her hands not visible, \textcolor[rgb]{1,0,0}{indicating a lack of tension or anxiety her overall demeanor suggests a sense of calmness and introspection, as she is engaging in self-reflection or admiring her own appearance, which is often associated with a feeling of surprise}

\textbf{Score:} 2

\textbf{Reason:} The predicted description does not align with the actual description of the character's emotional state. The actual description indicates a sense of surprise or suspicion, while the predicted description infers a somber or melancholic mood, sadness, or depression. The overlap between the predicted and actual emotional states is low, resulting in a lower score.

}
\\
\midrule
Omni-Emotion & {
\textbf{Output:} In the video, a young woman dressed in a pink dress stands in a cozy indoor setting. Her expression shifts from surprise to confusion, ultimately revealing a look of concern. Her brows are slightly furrowed, her eyes wide open, and her lips slightly parted, as if she is trying to comprehend or respond to some unexpected information. This emotional transition suggests that she may have received shocking news or encountered an unforeseen situation, leaving her feeling puzzled and worried. The entire process showcases her immediate reaction to external stimuli and her inner turmoil.

\textbf{Score}: 10

\textbf{Reason:} The predicted description accurately captures the emotional state of the character, reflecting her initial surprise, transition to confusion, and ultimate concern, aligning with the actual description of the character's emotional state of surprise or suspicion. The overlap between the predicted and actual emotional states is high, resulting in a higher score."
}  \\
\midrule

\end{tabular}
}
\label{table:case1}
\end{table*}

\begin{table*}[t]
\caption{Examples of multimodal emotion reasoning comparing Omni-Emotion with other emotion MLLMs. We highlight the incorrect reasoning with red color.}
\centering  
\scalebox{0.85}{
    \begin{tabular}{l|p{16cm} }
\toprule
\multicolumn{2}{l}{ The reasoning results of ``sample\_00000927.mp4'' In EMER~\cite{lian2023explainable} dataset} \\
\midrule
& {\includegraphics[height=2.3cm]{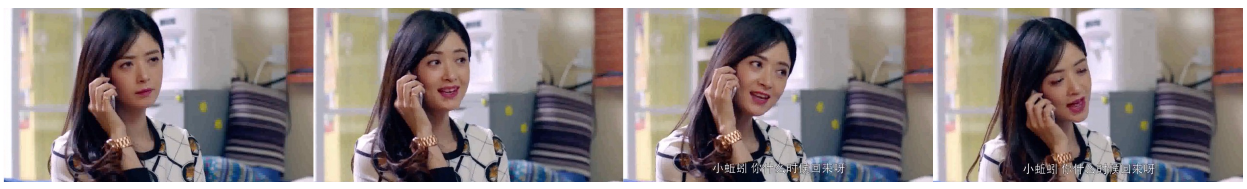}} \\
Subtitle (En) & {Little worm, when will you come back?} \\
\midrule
Ground Truth & {In the video, in the opening scene, we see a lady in an indoor environment. Her facial expression appears slightly serious, with a furrowed brow and tightly closed lips, as if she is attentively listening to a conversation on the phone. As time passes, in the middle scene, there is a noticeable change in the lady's expression. Her lips curve upwards, revealing a smile, and her eyes show relaxation and joy. This change may indicate that the conversation on the phone has become positive or that she has heard news that made her happy. The entire conversation process seems positive, making her feel happy. In the audio, the tone is gentle, with a slight rising intonation, and there is a sense of anticipation and joy in the speech. In the text, the subtitle says, ``Little worm, when will you come back?'' This sentence is likely what the lady said to the person on the phone. Based on the lady's facial expression change from seriousness to a smile in the video clues, as well as the description of a gentle tone with anticipation and joy in the audio clues, we can infer that this sentence is a greeting or expectation from the lady to an intimate person. The lady's expression and tone both show positive emotions, indicating her happiness and anticipation for the person's return.}  \\
\midrule
Emotion-LLaMA & {
\textbf{Output:} in the video, the woman's facial expressions and body language suggest a neutral emotional state her facial micro expressions, such as brow lowerer, lip corner depressor, and slight frown,  \textcolor[rgb]{1,0,0}{do not convey any strong emotions her tone and intonation are also neutral}, which further supports the inference of a neutral emotional state the fact that she is holding a cell phone and talking on it, possibly engaging in a conversation or checking her messages, suggests that she is focused on a task or activity rather than experiencing any intense emotions

\textbf{Score:} 3

\textbf{Reason:} The predicted description does not align with the actual description of the character's emotional state. The actual description indicates happiness and anticipation, while the predicted description infers a neutral emotional state. The overlap between the predicted and actual emotional states is low, resulting in a lower score.

}  \\
\midrule
Omni-Emotion & {
\textbf{Output:} In the video, the woman's emotional state is one of warmth and anticipation. She is wearing a white top with black polka dot accents and has long, dark hair. Her facial expression is gentle and her eyes are filled with warmth and hope. She is smiling and her body language is relaxed, indicating that she is enjoying the conversation on the other end of the phone. As the conversation progresses, her smile deepens, showing her inner joy and anticipation. When she mentions ``Xiao Qiu Yin, when are you coming back?'', her tone is calm and slightly excited, revealing her eagerness for the return of a loved one. Overall, her emotional state is one of warmth, happiness, and hopeful anticipation.

\textbf{Score:} 10

\textbf{Reason:}The predicted description accurately captures the emotional state of the character, reflecting her warmth, happiness, and hopeful anticipation, aligning with the actual description of the character's joyful and anticipatory emotional state. The overlap between the predicted and actual emotional states is high, resulting in a higher score.
}  \\

\midrule

\end{tabular}
}
\label{table:case3}
\end{table*}

\begin{table*}[t]
\caption{Examples of multimodal emotion reasoning comparing Omni-Emotion with other emotion MLLMs. We highlight the incorrect reasoning with red color.}
\centering  
\scalebox{0.85}{
    \begin{tabular}{l|p{16cm} }
\toprule
\multicolumn{2}{l}{ The reasoning results of ``sample\_00002292.mp4'' In EMER~\cite{lian2023explainable} dataset} \\
\midrule
& {\includegraphics[height=2.3cm]{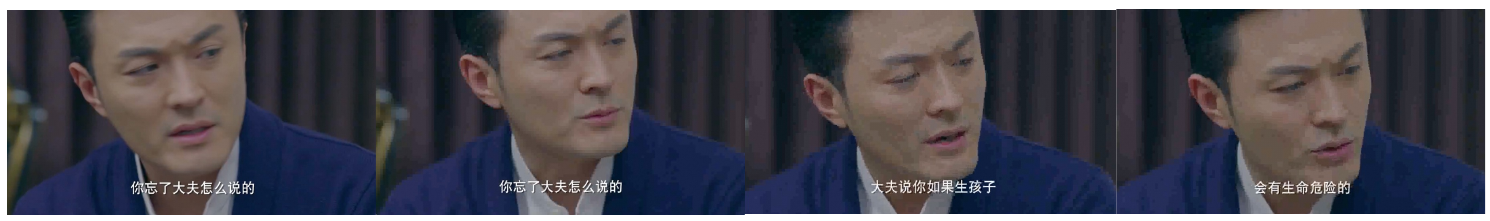}} \\
Subtitle (En) & {You forgot what the doctor said, the doctor said if you have a child, there will be a risk to your life} \\
\midrule
Ground Truth & {In the video, the screen shows a male character in what appears to be an indoor environment, as curtains can be seen in the background. In the video, he frowns and his facial expression looks serious, indicating some nervousness and concern. He maintains direct eye contact with the other person and his mouth moves, indicating that he is engaged in a serious conversation. Towards the end of the video, he accompanies his speech with head movements, which often signify emphasis. Overall, he is engaged in a conversation that is not easygoing. In the audio, the character's tone is serious. Combined with the text content, it seems to carry a sense of admonishment towards the other person. In the text, the subtitle reads: ``You forgot what the doctor said, the doctor said if you have a child, there will be a risk to your life.'' This sentence may be an expression of advice or warning from the male character to the other person. Based on the serious facial expression and direct eye contact of the male character in the video, as well as the emphasized head movements, it can be inferred that he is engaged in a serious conversation and his tone is serious. Combined with the description of the serious tone and admonishing meaning in the audio clues, it can be speculated that this sentence is the male character reminding the other person about the potential life-threatening risks of having a child, expressing his concern and nervousness.}  \\
\midrule
Emotion-LLaMA & {
\textbf{Output:} in the video, the man's facial expressions, including brow lowerer, upper lid raiser, and lip corner puller, suggest a state of anger his furrowed brow and tightened lips further reinforce this emotion his tone and intonation are also  \textcolor[rgb]{1,0,0}{angry}, which is evident from his raised voice and aggressive tone the physical cues, \textcolor[rgb]{1,0,0}{such as his clenched fists and tense posture, indicate that he is physically prepared for confrontation or aggression the combination of these facial expressions, tone, and physical cues suggests that the man is experiencing a strong negative emotion, likely anger, due to his intense focus on the object in front of him, which may be a threat or a source of frustration}

\textbf{Score:} 4

\textbf{Reason:} The predicted description only focuses on the emotional state of the character, which is anger, while the actual description indicates a serious and concerned emotional state. There is a mismatch between the predicted and actual emotional states, resulting in a lower score.

}  \\
\midrule
Omni-Emotion & {
\textbf{Output:} In the video, we can infer that the man is experiencing a complex emotional state. His facial expressions show a mix of confusion and deep contemplation, with furrowed brows and a slightly open mouth indicating his inner turmoil. The subtle changes in his facial muscles suggest that he is struggling to understand or process information. Additionally, his voice carries a tone of uncertainty and hesitation, with a slower pace and a slight tremor in his speech, reflecting his internal conflict and anxiety about the unknown future. Overall, the man appears to be grappling with a significant decision or dilemma, as evidenced by his intense focus and emotional fluctuations.

\textbf{Score:} 8

\textbf{Reason:} The predicted description accurately captures the emotional state of the character, reflecting his inner turmoil, anxiety, uncertainty, and deep contemplation, which aligns with the actual description of the character's nervousness, concern, and serious tone. The overlap between the predicted and actual emotional states is high, resulting in a higher score.
}  \\

\midrule

\end{tabular}
}
\label{table:case4}
\end{table*}

\begin{table*}[t]
\caption{Annotations comparison between our datasets, EMER, and MERR.}
\centering  
\scalebox{0.85}{
    \begin{tabular}{l|p{11cm}|l|p{3cm}}
\toprule

Dataset & Annotation & Number & Check\\

\midrule
EMER~\cite{lian2023explainable} & {
\textbf{One-hot label}: surprise

\textbf{EMER description}: In the video, the screen shows a male character in an indoor setting. At the beginning of the video, his eyes are wide open and his mouth is also open, indicating a surprised facial expression. In the following scenes, he looks around, seemingly explaining or narrating something to the people around him. Overall, his emotions are not positive or optimistic. In the audio, the character speaks with a stutter, which usually expresses feelings of nervousness, anxiety, or unease. Combined with the text content, the character seems to be unhappy and angry due to the prejudice of the people around him. The subtitle in the text says, "Why are you all looking at me like that? So, as long as it's a woman, does she have to have a relationship with me?" This sentence expresses the male character's dissatisfaction and anger towards the people around him. Based on the surprised and negative facial expression of the male character in the video clues, as well as the stuttering speech in the audio clues, we can infer that the male character is expressing a feeling of dissatisfaction and anger in this sentence. He may feel troubled by the prejudice of the people around him and is unhappy with this unfair treatment.

\textbf{Open-vocabulary labels}: surprise, nervous, dissatisfied
} & 332 & None\\
\midrule
MERR~\cite{cheng2024emotion} & {

\textbf{Coarse-grained annotation:}
The woman in the video is talking to a man, possibly discussing something important or sharing her thoughts and feelings. The woman’s expression and actions include eyes widened and mouth wide open, and the woman in the video speaks with an excited voice. Saying: Oh my!

\textbf{Fine-grained description:}
In the video, a woman is conversing with a man. Her facial expressions, with eyes widened and mouth wide open, clearly show surprise. She amplifies this visual cue with an excited tone as she exclaims, “Oh my.” This combination of voice and expression indicates that she is experiencing surprise, likely in response to unexpected news shared during the conversation.
}& 28,618/4,487 & None\\
\midrule
Ours & {

\textbf{Original label}: angry

\textbf{EMER description}: The scene in the video is set in a private space, with walls painted bright red, creating a warm and slightly oppressive atmosphere. The focus of the picture is on a woman, who is wearing a yellow top with a black watch as an embellishment. The woman's facial expression shows emotional fluctuations of surprise, confusion, struggle and finally anger, with her eyes wide open, brows furrowed, and mouth slightly open to show teeth. From the video, we can clearly feel the rapid emotional evolution of the woman's heart, which seems to be an emotional reaction of psychological tension and anger caused by receiving unexpected news. According to voice recognition technology, her voice is accompanied by shouts of surprise or rapid breathing, which will echo her facial expressions and body language, further confirming that she is experiencing a strong emotional reaction. Comprehensive analysis shows that the woman in the video has experienced an emotional transition from surprise to obvious anger and dissatisfaction. Her facial expressions and body language clearly show this emotional journey, showing an immediate emotional reaction of being violated or dissatisfied.

\textbf{Open-vocabulary labels}: angry, surprised, dissatisfied

} & 24,137/3,500 &  {Self-review against 
the original labels
and human review.}\\
\midrule
\end{tabular}
}
\label{table:diff}
\end{table*}

{
    \small
    \bibliographystyle{ieeenat_fullname}
    \bibliography{main}

\begin{thebibliography}{67}
\providecommand{\natexlab}[1]{#1}
\providecommand{\url}[1]{\texttt{#1}}
\expandafter\ifx\csname urlstyle\endcsname\relax
  \providecommand{\doi}[1]{doi: #1}\else
  \providecommand{\doi}{doi: \begingroup \urlstyle{rm}\Url}\fi

\bibitem[Alayrac et~al.(2022)Alayrac, Donahue, Luc, Miech, Barr, Hasson, Lenc, Mensch, Millican, Reynolds, et~al.]{alayrac2022flamingo}
Jean-Baptiste Alayrac, Jeff Donahue, Pauline Luc, Antoine Miech, Iain Barr, Yana Hasson, Karel Lenc, Arthur Mensch, Katherine Millican, Malcolm Reynolds, et~al.
\newblock Flamingo: a visual language model for few-shot learning.
\newblock \emph{Advances in neural information processing systems}, 35:\penalty0 23716--23736, 2022.

\bibitem[Ardila et~al.(2019)Ardila, Branson, Davis, Henretty, Kohler, Meyer, Morais, Saunders, Tyers, and Weber]{ardila2019common}
Rosana Ardila, Megan Branson, Kelly Davis, Michael Henretty, Michael Kohler, Josh Meyer, Reuben Morais, Lindsay Saunders, Francis~M Tyers, and Gregor Weber.
\newblock Common voice: A massively-multilingual speech corpus.
\newblock \emph{arXiv preprint arXiv:1912.06670}, 2019.

\bibitem[Baevski et~al.(2020)Baevski, Zhou, Mohamed, and Auli]{baevski2020wav2vec}
Alexei Baevski, Yuhao Zhou, Abdelrahman Mohamed, and Michael Auli.
\newblock wav2vec 2.0: A framework for self-supervised learning of speech representations.
\newblock \emph{Advances in neural information processing systems}, 33:\penalty0 12449--12460, 2020.

\bibitem[Bai et~al.(2023)Bai, Bai, Yang, Wang, Tan, Wang, Lin, Zhou, and Zhou]{qwen}
Jinze Bai, Shuai Bai, Shusheng Yang, Shijie Wang, Sinan Tan, Peng Wang, Junyang Lin, Chang Zhou, and Jingren Zhou.
\newblock Qwen-vl: A frontier large vision-language model with versatile abilities.
\newblock \emph{arXiv preprint arXiv:2308.12966}, 2023.

\bibitem[Chen et~al.(2020)Chen, Xie, Vedaldi, and Zisserman]{chen2020vggsound}
Honglie Chen, Weidi Xie, Andrea Vedaldi, and Andrew Zisserman.
\newblock Vggsound: A large-scale audio-visual dataset.
\newblock In \emph{ICASSP 2020-2020 IEEE International Conference on Acoustics, Speech and Signal Processing (ICASSP)}, pages 721--725. IEEE, 2020.

\bibitem[Chen et~al.(2024{\natexlab{a}})Chen, Huang, Dong, Zheng, and Shao]{chen2024finecliper}
Haodong Chen, Haojian Huang, Junhao Dong, Mingzhe Zheng, and Dian Shao.
\newblock Finecliper: Multi-modal fine-grained clip for dynamic facial expression recognition with adapters.
\newblock In \emph{Proceedings of the 32nd ACM International Conference on Multimedia}, pages 2301--2310, 2024{\natexlab{a}}.

\bibitem[Chen et~al.(2023{\natexlab{a}})Chen, Zhu, Shen, Li, Liu, Zhang, Krishnamoorthi, Chandra, Xiong, and Elhoseiny]{chen2023minigpt}
Jun Chen, Deyao Zhu, Xiaoqian Shen, Xiang Li, Zechun Liu, Pengchuan Zhang, Raghuraman Krishnamoorthi, Vikas Chandra, Yunyang Xiong, and Mohamed Elhoseiny.
\newblock Minigpt-v2: large language model as a unified interface for vision-language multi-task learning.
\newblock \emph{arXiv preprint arXiv:2310.09478}, 2023{\natexlab{a}}.

\bibitem[Chen et~al.(2023{\natexlab{b}})Chen, Zhang, Zeng, Zhang, Zhu, and Zhao]{shikra}
Keqin Chen, Zhao Zhang, Weili Zeng, Richong Zhang, Feng Zhu, and Rui Zhao.
\newblock Shikra: Unleashing multimodal llm's referential dialogue magic.
\newblock \emph{arXiv preprint arXiv:2306.15195}, 2023{\natexlab{b}}.

\bibitem[Chen et~al.(2024{\natexlab{b}})Chen, Wang, Cao, Liu, Gao, Cui, Zhu, Ye, Tian, Liu, et~al.]{chen2024expanding}
Zhe Chen, Weiyun Wang, Yue Cao, Yangzhou Liu, Zhangwei Gao, Erfei Cui, Jinguo Zhu, Shenglong Ye, Hao Tian, Zhaoyang Liu, et~al.
\newblock Expanding performance boundaries of open-source multimodal models with model, data, and test-time scaling.
\newblock \emph{arXiv preprint arXiv:2412.05271}, 2024{\natexlab{b}}.

\bibitem[Cheng et~al.(2024)Cheng, Cheng, He, Sun, Wang, Lin, Lian, Peng, and Hauptmann]{cheng2024emotion}
Zebang Cheng, Zhi-Qi Cheng, Jun-Yan He, Jingdong Sun, Kai Wang, Yuxiang Lin, Zheng Lian, Xiaojiang Peng, and Alexander Hauptmann.
\newblock Emotion-llama: Multimodal emotion recognition and reasoning with instruction tuning.
\newblock \emph{arXiv preprint arXiv:2406.11161}, 2024.

\bibitem[Chiang et~al.(2023)Chiang, Li, Lin, Sheng, Wu, Zhang, Zheng, Zhuang, Zhuang, Gonzalez, Stoica, and Xing]{vicuna2023}
Wei-Lin Chiang, Zhuohan Li, Zi Lin, Ying Sheng, Zhanghao Wu, Hao Zhang, Lianmin Zheng, Siyuan Zhuang, Yonghao Zhuang, Joseph~E. Gonzalez, Ion Stoica, and Eric~P. Xing.
\newblock Vicuna: An open-source chatbot impressing gpt-4 with 90\%* chatgpt quality, 2023.

\bibitem[Chu et~al.(2024)Chu, Xu, Yang, Wei, Wei, Guo, Leng, Lv, He, Lin, Zhou, and Zhou]{Qwen2Audio}
Yunfei Chu, Jin Xu, Qian Yang, Haojie Wei, Xipin Wei, Zhifang Guo, Yichong Leng, Yuanjun Lv, Jinzheng He, Junyang Lin, Chang Zhou, and Jingren Zhou.
\newblock Qwen2-audio technical report.
\newblock \emph{arXiv preprint arXiv:2407.10759}, 2024.

\bibitem[Chumachenko et~al.(2024)Chumachenko, Iosifidis, and Gabbouj]{chumachenko2024mma}
Kateryna Chumachenko, Alexandros Iosifidis, and Moncef Gabbouj.
\newblock Mma-dfer: Multimodal adaptation of unimodal models for dynamic facial expression recognition in-the-wild.
\newblock In \emph{Proceedings of the IEEE/CVF Conference on Computer Vision and Pattern Recognition}, pages 4673--4682, 2024.

\bibitem[Deng et~al.(2019)Deng, Guo, Zhou, Yu, Kotsia, and Zafeiriou]{deng2019retinaface}
Jiankang Deng, Jia Guo, Yuxiang Zhou, Jinke Yu, Irene Kotsia, and Stefanos Zafeiriou.
\newblock Retinaface: Single-stage dense face localisation in the wild.
\newblock \emph{arXiv preprint arXiv:1905.00641}, 2019.

\bibitem[Devlin et~al.(2018)Devlin, Chang, Lee, and Toutanova]{devlin2018bert}
Jacob Devlin, Ming-Wei Chang, Kenton Lee, and Kristina Toutanova.
\newblock Bert: Pre-training of deep bidirectional transformers for language understanding.
\newblock \emph{arXiv preprint arXiv:1810.04805}, 2018.

\bibitem[Fan et~al.(2021)Fan, Xu, Xing, Chen, and Huang]{fan2021lssed}
Weiquan Fan, Xiangmin Xu, Xiaofen Xing, Weidong Chen, and Dongyan Huang.
\newblock Lssed: a large-scale dataset and benchmark for speech emotion recognition.
\newblock In \emph{ICASSP 2021-2021 IEEE International Conference on Acoustics, Speech and Signal Processing (ICASSP)}, pages 641--645. IEEE, 2021.

\bibitem[Hsu et~al.(2021)Hsu, Bolte, Tsai, Lakhotia, Salakhutdinov, and Mohamed]{hsu2021hubert}
Wei-Ning Hsu, Benjamin Bolte, Yao-Hung~Hubert Tsai, Kushal Lakhotia, Ruslan Salakhutdinov, and Abdelrahman Mohamed.
\newblock Hubert: Self-supervised speech representation learning by masked prediction of hidden units.
\newblock \emph{IEEE/ACM Transactions on Audio, Speech, and Language Processing}, 29:\penalty0 3451--3460, 2021.

\bibitem[Hung and Alias(2023)]{hung2023beyond}
Lai~Po Hung and Suraya Alias.
\newblock Beyond sentiment analysis: A review of recent trends in text based sentiment analysis and emotion detection.
\newblock \emph{Journal of Advanced Computational Intelligence and Intelligent Informatics}, 27\penalty0 (1):\penalty0 84--95, 2023.

\bibitem[Jiang et~al.(2020)Jiang, Zong, Zheng, Tang, Xia, Lu, and Liu]{jiang2020dfew}
Xingxun Jiang, Yuan Zong, Wenming Zheng, Chuangao Tang, Wanchuang Xia, Cheng Lu, and Jiateng Liu.
\newblock Dfew: A large-scale database for recognizing dynamic facial expressions in the wild.
\newblock In \emph{Proceedings of the 28th ACM international conference on multimedia}, pages 2881--2889, 2020.

\bibitem[Kondratenko et~al.(2022)Kondratenko, Sokolov, Karpov, Kutuzov, Savushkin, and Minkin]{kondratenko2022large}
Vladimir Kondratenko, Artem Sokolov, Nikolay Karpov, Oleg Kutuzov, Nikita Savushkin, and Fyodor Minkin.
\newblock Large raw emotional dataset with aggregation mechanism.
\newblock \emph{arXiv preprint arXiv:2212.12266}, 2022.

\bibitem[Kossaifi et~al.(2017)Kossaifi, Tzimiropoulos, Todorovic, and Pantic]{kossaifi2017afew}
Jean Kossaifi, Georgios Tzimiropoulos, Sinisa Todorovic, and Maja Pantic.
\newblock Afew-va database for valence and arousal estimation in-the-wild.
\newblock \emph{Image and Vision Computing}, 65:\penalty0 23--36, 2017.

\bibitem[Kuprashevich and Tolstykh(2023)]{kuprashevich2023mivolo}
Maksim Kuprashevich and Irina Tolstykh.
\newblock Mivolo: Multi-input transformer for age and gender estimation.
\newblock In \emph{International Conference on Analysis of Images, Social Networks and Texts}, pages 212--226. Springer, 2023.

\bibitem[Lee et~al.(2019)Lee, Kim, Kim, Park, and Sohn]{lee2019context}
Jiyoung Lee, Seungryong Kim, Sunok Kim, Jungin Park, and Kwanghoon Sohn.
\newblock Context-aware emotion recognition networks.
\newblock In \emph{Proceedings of the IEEE/CVF international conference on computer vision}, pages 10143--10152, 2019.

\bibitem[Lei et~al.(2023)Lei, Dong, Wang, Wang, and Wang]{lei2023instructerc}
Shanglin Lei, Guanting Dong, Xiaoping Wang, Keheng Wang, and Sirui Wang.
\newblock Instructerc: Reforming emotion recognition in conversation with a retrieval multi-task llms framework.
\newblock \emph{arXiv preprint arXiv:2309.11911}, 2023.

\bibitem[Li et~al.(2024{\natexlab{a}})Li, Zhang, Guo, Zhang, Li, Zhang, Zhang, Li, Liu, and Li]{li2024llava}
Bo Li, Yuanhan Zhang, Dong Guo, Renrui Zhang, Feng Li, Hao Zhang, Kaichen Zhang, Yanwei Li, Ziwei Liu, and Chunyuan Li.
\newblock Llava-onevision: Easy visual task transfer.
\newblock \emph{arXiv preprint arXiv:2408.03326}, 2024{\natexlab{a}}.

\bibitem[Li et~al.(2023{\natexlab{a}})Li, Li, Savarese, and Hoi]{li2023blip}
Junnan Li, Dongxu Li, Silvio Savarese, and Steven Hoi.
\newblock Blip-2: Bootstrapping language-image pre-training with frozen image encoders and large language models.
\newblock In \emph{International Conference on Machine Learning}, pages 1--13, 2023{\natexlab{a}}.

\bibitem[Li et~al.(2023{\natexlab{b}})Li, He, Wang, Li, Wang, Luo, Wang, Wang, and Qiao]{li2023videochat}
KunChang Li, Yinan He, Yi Wang, Yizhuo Li, Wenhai Wang, Ping Luo, Yali Wang, Limin Wang, and Yu Qiao.
\newblock Videochat: Chat-centric video understanding.
\newblock \emph{arXiv preprint arXiv:2305.06355}, 2023{\natexlab{b}}.

\bibitem[Li et~al.(2024{\natexlab{b}})Li, Wang, He, Li, Wang, Liu, Wang, Xu, Chen, Luo, Wang, and Qiao]{li2024mvbench}
Kunchang Li, Yali Wang, Yinan He, Yizhuo Li, Yi Wang, Yi Liu, Zun Wang, Jilan Xu, Guo Chen, Ping Luo, Limin Wang, and Yu Qiao.
\newblock Mvbench: A comprehensive multi-modal video understanding benchmark.
\newblock In \emph{Proceedings of the IEEE/CVF Conference on Computer Vision and Pattern Recognition}, 2024{\natexlab{b}}.

\bibitem[Li et~al.(2023{\natexlab{c}})Li, Wang, and Cui]{li2023decoupled}
Yong Li, Yuanzhi Wang, and Zhen Cui.
\newblock Decoupled multimodal distilling for emotion recognition.
\newblock In \emph{Proceedings of the IEEE/CVF Conference on Computer Vision and Pattern Recognition}, pages 6631--6640, 2023{\natexlab{c}}.

\bibitem[Lian et~al.(2023)Lian, Sun, Xu, Sun, Xu, Wen, Chen, Liu, and Tao]{lian2023explainable}
Zheng Lian, Licai Sun, Mingyu Xu, Haiyang Sun, Ke Xu, Zhuofan Wen, Shun Chen, Bin Liu, and Jianhua Tao.
\newblock Explainable multimodal emotion reasoning.
\newblock \emph{arXiv preprint arXiv:2306.15401}, 2023.

\bibitem[Lian et~al.(2024{\natexlab{a}})Lian, Sun, Sun, Wen, Zhang, Chen, Gu, Zhao, Ma, Chen, et~al.]{lian2024mer}
Zheng Lian, Haiyang Sun, Licai Sun, Zhuofan Wen, Siyuan Zhang, Shun Chen, Hao Gu, Jinming Zhao, Ziyang Ma, Xie Chen, et~al.
\newblock Mer 2024: Semi-supervised learning, noise robustness, and open-vocabulary multimodal emotion recognition.
\newblock \emph{arXiv preprint arXiv:2404.17113}, 2024{\natexlab{a}}.

\bibitem[Lian et~al.(2024{\natexlab{b}})Lian, Sun, Sun, Yi, Liu, and Tao]{lian2024affectgpt}
Zheng Lian, Haiyang Sun, Licai Sun, Jiangyan Yi, Bin Liu, and Jianhua Tao.
\newblock Affectgpt: Dataset and framework for explainable multimodal emotion recognition.
\newblock \emph{arXiv preprint arXiv:2407.07653}, 2024{\natexlab{b}}.

\bibitem[Liu et~al.(2024)Liu, Li, Wu, and Lee]{liu2024visual}
Haotian Liu, Chunyuan Li, Qingyang Wu, and Yong~Jae Lee.
\newblock Visual instruction tuning.
\newblock \emph{NeurIPS}, 36, 2024.

\bibitem[Liu et~al.(2022)Liu, Dai, Feng, Wang, Yin, Zeng, and Shan]{liu2022mafw}
Yuanyuan Liu, Wei Dai, Chuanxu Feng, Wenbin Wang, Guanghao Yin, Jiabei Zeng, and Shiguang Shan.
\newblock Mafw: A large-scale, multi-modal, compound affective database for dynamic facial expression recognition in the wild.
\newblock In \emph{Proceedings of the 30th ACM International Conference on Multimedia}, pages 24--32, 2022.

\bibitem[Livingstone and Russo(2018)]{livingstone2018ryerson}
Steven~R Livingstone and Frank~A Russo.
\newblock The ryerson audio-visual database of emotional speech and song (ravdess): A dynamic, multimodal set of facial and vocal expressions in north american english.
\newblock \emph{PloS One}, 13\penalty0 (5):\penalty0 e0196391, 2018.

\bibitem[Luo et~al.(2023)Luo, Zhao, Yang, Dong, Qiu, Lu, Wang, and Wei]{luo2023valley}
Ruipu Luo, Ziwang Zhao, Min Yang, Junwei Dong, Minghui Qiu, Pengcheng Lu, Tao Wang, and Zhongyu Wei.
\newblock Valley: Video assistant with large language model enhanced ability.
\newblock \emph{arXiv preprint arXiv:2306.07207}, 2023.

\bibitem[Maaz et~al.(2023)Maaz, Rasheed, Khan, and Khan]{maaz2023video}
Muhammad Maaz, Hanoona Rasheed, Salman Khan, and Fahad~Shahbaz Khan.
\newblock Video-chatgpt: Towards detailed video understanding via large vision and language models.
\newblock \emph{arXiv preprint arXiv:2306.05424}, 2023.

\bibitem[Mei et~al.(2024)Mei, Meng, Liu, Kong, Ko, Zhao, Plumbley, Zou, and Wang]{mei2024wavcaps}
Xinhao Mei, Chutong Meng, Haohe Liu, Qiuqiang Kong, Tom Ko, Chengqi Zhao, Mark~D Plumbley, Yuexian Zou, and Wenwu Wang.
\newblock Wavcaps: A chatgpt-assisted weakly-labelled audio captioning dataset for audio-language multimodal research.
\newblock \emph{IEEE/ACM Transactions on Audio, Speech, and Language Processing}, 2024.

\bibitem[Narayan et~al.(2024)Narayan, VS, Chellappa, and Patel]{narayan2024facexformer}
Kartik Narayan, Vibashan VS, Rama Chellappa, and Vishal~M Patel.
\newblock Facexformer: A unified transformer for facial analysis.
\newblock \emph{arXiv preprint arXiv:2403.12960}, 2024.

\bibitem[Ngwe et~al.(2023)Ngwe, Lim, Lee, and Ong]{ngwe2023patt}
Jia~Le Ngwe, Kian~Ming Lim, Chin~Poo Lee, and Thian~Song Ong.
\newblock Patt-lite: Lightweight patch and attention mobilenet for challenging facial expression recognition.
\newblock \emph{arXiv preprint arXiv:2306.09626}, 2023.

\bibitem[OpenAI(2023{\natexlab{a}})]{openai2023chatgpt}
OpenAI.
\newblock Chatgpt.
\newblock https://openai.com/blog/chatgpt/, 2023{\natexlab{a}}.

\bibitem[OpenAI(2023{\natexlab{b}})]{openai2023gpt4v}
OpenAI.
\newblock Gpt-4v(ision) system card, 2023{\natexlab{b}}.

\bibitem[Ouyang et~al.(2022)Ouyang, Wu, Jiang, Almeida, Wainwright, Mishkin, Zhang, Agarwal, Slama, Ray, et~al.]{ouyang2022instruct-tuning}
Long Ouyang, Jeffrey Wu, Xu Jiang, Diogo Almeida, Carroll Wainwright, Pamela Mishkin, Chong Zhang, Sandhini Agarwal, Katarina Slama, Alex Ray, et~al.
\newblock Training language models to follow instructions with human feedback.
\newblock \emph{Advances in Neural Information Processing Systems}, 2022.

\bibitem[Peng et~al.(2023)Peng, Wang, Dong, Hao, Huang, Ma, and Wei]{peng2023kosmos}
Zhiliang Peng, Wenhui Wang, Li Dong, Yaru Hao, Shaohan Huang, Shuming Ma, and Furu Wei.
\newblock Kosmos-2: Grounding multimodal large language models to the world.
\newblock \emph{arXiv preprint arXiv:2306.14824}, 2023.

\bibitem[Radford et~al.(2021)Radford, Kim, Hallacy, Ramesh, Goh, Agarwal, Sastry, Askell, Mishkin, Clark, et~al.]{radford2021learning}
Alec Radford, Jong~Wook Kim, Chris Hallacy, Aditya Ramesh, Gabriel Goh, Sandhini Agarwal, Girish Sastry, Amanda Askell, Pamela Mishkin, Jack Clark, et~al.
\newblock Learning transferable visual models from natural language supervision.
\newblock In \emph{International conference on machine learning}, pages 8748--8763. PMLR, 2021.

\bibitem[Radford et~al.(2023)Radford, Kim, Xu, Brockman, McLeavey, and Sutskever]{radford2023robust}
Alec Radford, Jong~Wook Kim, Tao Xu, Greg Brockman, Christine McLeavey, and Ilya Sutskever.
\newblock Robust speech recognition via large-scale weak supervision.
\newblock In \emph{International conference on machine learning}, pages 28492--28518. PMLR, 2023.

\bibitem[Su et~al.(2023)Su, Lan, Li, Xu, Wang, and Cai]{su2023pandagpt}
Yixuan Su, Tian Lan, Huayang Li, Jialu Xu, Yan Wang, and Deng Cai.
\newblock Pandagpt: One model to instruction-follow them all.
\newblock In \emph{Proceedings of the 1st Workshop on Taming Large Language Models: Controllability in the era of Interactive Assistants}, pages 11--23, 2023.

\bibitem[Sun et~al.(2023)Sun, Lian, Liu, and Tao]{sun2023mae}
Licai Sun, Zheng Lian, Bin Liu, and Jianhua Tao.
\newblock Mae-dfer: Efficient masked autoencoder for self-supervised dynamic facial expression recognition.
\newblock In \emph{Proceedings of the 31st ACM International Conference on Multimedia}, pages 6110--6121, 2023.

\bibitem[Sun et~al.(2024{\natexlab{a}})Sun, Lian, Liu, and Tao]{sun2024hicmae}
Licai Sun, Zheng Lian, Bin Liu, and Jianhua Tao.
\newblock Hicmae: Hierarchical contrastive masked autoencoder for self-supervised audio-visual emotion recognition.
\newblock \emph{Information Fusion}, 108:\penalty0 102382, 2024{\natexlab{a}}.

\bibitem[Sun et~al.(2024{\natexlab{b}})Sun, Lian, Wang, He, Xu, Sun, Liu, and Tao]{sun2024svfap}
Licai Sun, Zheng Lian, Kexin Wang, Yu He, Mingyu Xu, Haiyang Sun, Bin Liu, and Jianhua Tao.
\newblock Svfap: Self-supervised video facial affect perceiver.
\newblock \emph{IEEE Transactions on Affective Computing}, 2024{\natexlab{b}}.

\bibitem[Tong et~al.(2024)Tong, Brown, Wu, Woo, Middepogu, Akula, Yang, Yang, Iyer, Pan, et~al.]{tong2024cambrian}
Shengbang Tong, Ellis Brown, Penghao Wu, Sanghyun Woo, Manoj Middepogu, Sai~Charitha Akula, Jihan Yang, Shusheng Yang, Adithya Iyer, Xichen Pan, et~al.
\newblock Cambrian-1: A fully open, vision-centric exploration of multimodal llms.
\newblock \emph{arXiv preprint arXiv:2406.16860}, 2024.

\bibitem[Wang et~al.(2020)Wang, Peng, Yang, Lu, and Qiao]{wang2020suppressing}
Kai Wang, Xiaojiang Peng, Jianfei Yang, Shijian Lu, and Yu Qiao.
\newblock Suppressing uncertainties for large-scale facial expression recognition.
\newblock In \emph{Proceedings of the IEEE/CVF conference on computer vision and pattern recognition}, pages 6897--6906, 2020.

\bibitem[Wang et~al.(2024{\natexlab{a}})Wang, Bai, Tan, Wang, Fan, Bai, Chen, Liu, Wang, Ge, et~al.]{wang2024qwen2}
Peng Wang, Shuai Bai, Sinan Tan, Shijie Wang, Zhihao Fan, Jinze Bai, Keqin Chen, Xuejing Liu, Jialin Wang, Wenbin Ge, et~al.
\newblock Qwen2-vl: Enhancing vision-language model's perception of the world at any resolution.
\newblock \emph{arXiv preprint arXiv:2409.12191}, 2024{\natexlab{a}}.

\bibitem[Wang et~al.(2023)Wang, Chen, Chen, Wu, Zhu, Zeng, Luo, Lu, Zhou, Qiao, et~al.]{wang2023visionllm}
Wenhai Wang, Zhe Chen, Xiaokang Chen, Jiannan Wu, Xizhou Zhu, Gang Zeng, Ping Luo, Tong Lu, Jie Zhou, Yu Qiao, et~al.
\newblock Visionllm: Large language model is also an open-ended decoder for vision-centric tasks.
\newblock \emph{arXiv preprint arXiv:2305.11175}, 2023.

\bibitem[Wang et~al.(2022)Wang, Sun, Huang, Liu, Gao, Zhang, Ge, and Zhang]{wang2022ferv39k}
Yan Wang, Yixuan Sun, Yiwen Huang, Zhongying Liu, Shuyong Gao, Wei Zhang, Weifeng Ge, and Wenqiang Zhang.
\newblock Ferv39k: A large-scale multi-scene dataset for facial expression recognition in videos.
\newblock In \emph{Proceedings of the IEEE/CVF conference on computer vision and pattern recognition}, pages 20922--20931, 2022.

\bibitem[Wang et~al.(2024{\natexlab{b}})Wang, Li, Li, Yu, He, Chen, Pei, Zheng, Xu, Wang, et~al.]{wang2024internvideo2}
Yi Wang, Kunchang Li, Xinhao Li, Jiashuo Yu, Yinan He, Guo Chen, Baoqi Pei, Rongkun Zheng, Jilan Xu, Zun Wang, et~al.
\newblock Internvideo2: Scaling video foundation models for multimodal video understanding.
\newblock \emph{arXiv preprint arXiv:2403.15377}, 2024{\natexlab{b}}.

\bibitem[Wang et~al.(2024{\natexlab{c}})Wang, Li, and Cui]{wang2024incomplete}
Yuanzhi Wang, Yong Li, and Zhen Cui.
\newblock Incomplete multimodality-diffused emotion recognition.
\newblock \emph{Advances in Neural Information Processing Systems}, 36, 2024{\natexlab{c}}.

\bibitem[Wojke et~al.(2017)Wojke, Bewley, and Paulus]{wojke2017simple}
Nicolai Wojke, Alex Bewley, and Dietrich Paulus.
\newblock Simple online and realtime tracking with a deep association metric.
\newblock In \emph{2017 IEEE international conference on image processing (ICIP)}, pages 3645--3649. IEEE, 2017.

\bibitem[Xie et~al.(2024)Xie, Peng, Tseng, Chen, Hsu, Shuai, and Cheng]{xie2024emovit}
Hongxia Xie, Chu-Jun Peng, Yu-Wen Tseng, Hung-Jen Chen, Chan-Feng Hsu, Hong-Han Shuai, and Wen-Huang Cheng.
\newblock Emovit: Revolutionizing emotion insights with visual instruction tuning.
\newblock \emph{arXiv preprint arXiv:2404.16670}, 2024.

\bibitem[Zadeh et~al.(2018)Zadeh, Liang, Poria, Cambria, and Morency]{zadeh2018multimodal}
AmirAli~Bagher Zadeh, Paul~Pu Liang, Soujanya Poria, Erik Cambria, and Louis-Philippe Morency.
\newblock Multimodal language analysis in the wild: Cmu-mosei dataset and interpretable dynamic fusion graph.
\newblock In \emph{Proceedings of the 56th Annual Meeting of the Association for Computational Linguistics (Volume 1: Long Papers)}, pages 2236--2246, 2018.

\bibitem[Zhai et~al.(2023)Zhai, Mustafa, Kolesnikov, and Beyer]{zhai2023sigmoid}
Xiaohua Zhai, Basil Mustafa, Alexander Kolesnikov, and Lucas Beyer.
\newblock Sigmoid loss for language image pre-training.
\newblock In \emph{Proceedings of the IEEE/CVF International Conference on Computer Vision}, pages 11975--11986, 2023.

\bibitem[Zhang et~al.(2023{\natexlab{a}})Zhang, Li, and Bing]{zhang2023video}
Hang Zhang, Xin Li, and Lidong Bing.
\newblock Video-llama: An instruction-tuned audio-visual language model for video understanding.
\newblock In \emph{Proceedings of the Conference on Empirical Methods in Natural Language Processing: System Demonstrations}, pages 543--553, 2023{\natexlab{a}}.

\bibitem[Zhang et~al.(2023{\natexlab{b}})Zhang, Pan, and Wang]{zhang2023learning}
Sitao Zhang, Yimu Pan, and James~Z Wang.
\newblock Learning emotion representations from verbal and nonverbal communication.
\newblock In \emph{Proceedings of the IEEE/CVF Conference on Computer Vision and Pattern Recognition}, pages 18993--19004, 2023{\natexlab{b}}.

\bibitem[Zhang et~al.(2023{\natexlab{c}})Zhang, Li, Lin, Xu, and Xiao]{zhang2023transformer}
Xiaoqin Zhang, Min Li, Sheng Lin, Hang Xu, and Guobao Xiao.
\newblock Transformer-based multimodal emotional perception for dynamic facial expression recognition in the wild.
\newblock \emph{IEEE Transactions on Circuits and Systems for Video Technology}, 2023{\natexlab{c}}.

\bibitem[Zhao and Patras(2023)]{zhao2023prompting}
Zengqun Zhao and Ioannis Patras.
\newblock Prompting visual-language models for dynamic facial expression recognition.
\newblock \emph{arXiv preprint arXiv:2308.13382}, 2023.

\bibitem[Zhou et~al.(2019)Zhou, Meng, Zhang, Peng, Du, Wang, and Qiao]{zhou2019exploring}
Hengshun Zhou, Debin Meng, Yuanyuan Zhang, Xiaojiang Peng, Jun Du, Kai Wang, and Yu Qiao.
\newblock Exploring emotion features and fusion strategies for audio-video emotion recognition.
\newblock In \emph{2019 International conference on multimodal interaction}, pages 562--566, 2019.

\bibitem[Zhu et~al.(2023)Zhu, Chen, Shen, Li, and Elhoseiny]{zhu2023minigpt}
Deyao Zhu, Jun Chen, Xiaoqian Shen, Xiang Li, and Mohamed Elhoseiny.
\newblock Minigpt-4: Enhancing vision-language understanding with advanced large language models.
\newblock \emph{arXiv preprint arXiv:2304.10592}, 2023.

\end{thebibliography}
}

\end{document}